\documentclass[lettersize,journal]{IEEEtran}
\usepackage{amsmath,amsfonts}
\usepackage{algorithmic}
\usepackage{array}
\usepackage[caption=false,font=normalsize,labelfont=sf,textfont=sf]{subfig}
\usepackage{textcomp}
\usepackage{stfloats}
\usepackage{url}
\usepackage{verbatim}
\usepackage{graphicx}
\usepackage{makecell}
\usepackage{graphicx}
\usepackage{textcomp}
\usepackage{array} 
\usepackage{longtable}
\usepackage{booktabs}
\usepackage{float}
\usepackage{soul}
\usepackage{colortbl}
\usepackage{amssymb}
\usepackage{lineno}
\usepackage{graphicx}
\usepackage{cite}

\makeatletter
\newcommand*\bigcdot{\mathpalette\bigcdot@{.5}}
\newcommand*\bigcdot@[2]{\mathbin{\vcenter{\hbox{\scalebox{#2}{$\m@th#1\bullet$}}}}}
\makeatother
\def\BibTeX{{\rm B\kern-.05em{\sc i\kern-.025em b}\kern-.08em
		T\kern-.1667em\lower.7ex\hbox{E}\kern-.125emX}}
\usepackage{balance}
\begin{document}
	\setlength{\textfloatsep}{2pt}
	\title{Rmote Sensing Image Change Detection \\ With Graph Interaction}
	\author{Chenglong Liu}
	
	\maketitle
	
	\begin{abstract}
	Modern remote sensing image change detection (CD) has witnessed substantial advancements by harnessing the potent feature extraction capabilities of CNNs and Transforms.
	Yet, prevailing change detection techniques consistently prioritize extracting semantic features related to significant alterations, overlooking the viability of directly
	 interacting with bitemporal image features.In this letter, we propose a bitemporal image graph Interaction network  for remote sensing change detection, namely BGINet-CD.
	 More specifically, by leveraging the concept of non-local operations and mapping the features obtained from the backbone network to the graph structure space, we propose a 
	 unified self-focus mechanism for bitemporal images. This approach enhances the information coupling between the two temporal images while effectively suppressing task-irrelevant 
	 interference.Based on a streamlined backbone architecture, namely ResNet18, our model demonstrates superior performance compared to other state-of-the-art methods (SOTA) on the GZ CD dataset. 
	 Moreover, the model exhibits an enhanced trade-off between accuracy and computational efficiency, further improving its overall effectiveness.
	\end{abstract}
	
	
	\section{Introduction}
	\IEEEPARstart 
	{T}{HE} change detection is an important research topic in remote sensing, 
	as it aims to identify changes that occur between two images acquired at different times 
	in the same geographical location. With the growing availability and utilization of remote sensing 
	satellites, change detection has found widespread application in various fields. It is commonly 
	used for monitoring urban sprawl\cite{frick2019framework}, assessing damage caused by natural disasters\cite{A1999Change}, and conducting 
	surveys of urban and rural areas\cite{MSCANet}.Multi-temporal remote sensing images often contain a variety 
	of interferences due to different imaging conditions and shooting times. These interferences 
	include spectral differences caused by varying light intensity and seasonal changes, as well
	as differences in shooting angles that result in varying shapes of buildings within the scene.
    Consequently, these factors can introduce pseudo-change during the detection process.

		A strong model should accurately identify unrelated disturbances in diachronic images and distinguish natural changes from complex uncorrelated ones\cite{BIT}. 
	Existing methods for change detection can be broadly categorized into two main groups: traditional change detection methods and deepl learning  methods.
	Traditional change detection methods encompass various approaches, including algebraic operation-based, transform-based, and classification-based methods. 
	Algebraic operation-based methods involve direct pixel-wise comparison in multi-temporal images and the selection of an appropriate threshold to classify 
	pixels as changed or unchanged. Image transformation techniques, such as principal component analysis (PCA)\cite{PCA} and change vector analysis\cite{CVA}, are also commonly used.
	On the other hand, machine learning-based methods, such as support vector machines, random forests, and kernel regression, have emerged as alternative approaches in recent years.

		Deep learning-based approaches have gained prominence due to their powerful nonlinear feature extraction capabilities. These approaches have witnessed the proposal of several 
	attention mechanisms, such as spatial attention\cite{cbam}, channel attention\cite{senet}, and self-attention\cite{vaswani2017attention}, aimed at obtaining improved feature representations.
    Chen et al. effectively modeled the context in the visual space-time domain by visually representing the high-level concept of interest change\cite{BIT}. Fang et al. addressed the exact 
	spatial location loss from continuous sampling by combining DenseNet and NestedUnet\cite{snunet}. Additionally, Chen et al. proposed a novel edge loss that enhances the network's attention to details like boundary regions and small regions\cite{rdpnet}.

	 	Although the methods above have shown promising results, none have explored the possibility of feature interaction between bi-temporal images prior to extracting different features. Drawing inspiration from non-local operations and DMINet \cite{DMINet},
	we have proposed a bi-temporal image graph Interaction network(BGINet) to facilitate feature interaction between bi-temporal images. This approach enhances the information coupling between bi-temporal images by extracting 
	bi-temporal features using a backbone network and routing them through a graph interaction module, effectively suppressing uncorrelated changes.

	 To demonstrate the effectiveness of our method, we have utilized a simple backbone network (ResNet18) in BGINet-CD. Initially, the features obtained from the backbone network are subjected to soft clustering, with each cluster being mapped to a vertex in the graph space. The Graph Interaction Module (GIM) captures the coupling relationship between the bi-temporal images, thus enhancing the information coupling. Finally, the clusters are reprojected to their original spatial coordinates.

		The contributions of this letter are mainly as follows:

	\begin{enumerate}
			\item We propose BGINet-CD, a graph convolutional neural network for remote sensing change detection, which effectively enhances the information coupling between diachronic images.
			\item The introduction of the GIM module enables the capture of the coupling relationship between biphasic images, enhancing information coupling and suppressing uncorrelated changes.
			\item We conducted quantitative and qualitative experiments on two datasets, our experiments demonstrated that our proposed BGINet-CD achieves a desirable balance between accuracy and efficiency, achieving state-of-the-art performance on the GZ dataset.Code is available at \url{https://github.com/JackLiu-97/BGINet.git}.
	\end{enumerate}
	\begin{figure*}[!t]
		\centering
		\vspace{-0.8cm}   
		\setlength{\abovecaptionskip}{0.1cm} 
		\setlength{\belowdisplayskip}{1pt} 
		\includegraphics[scale=0.3]{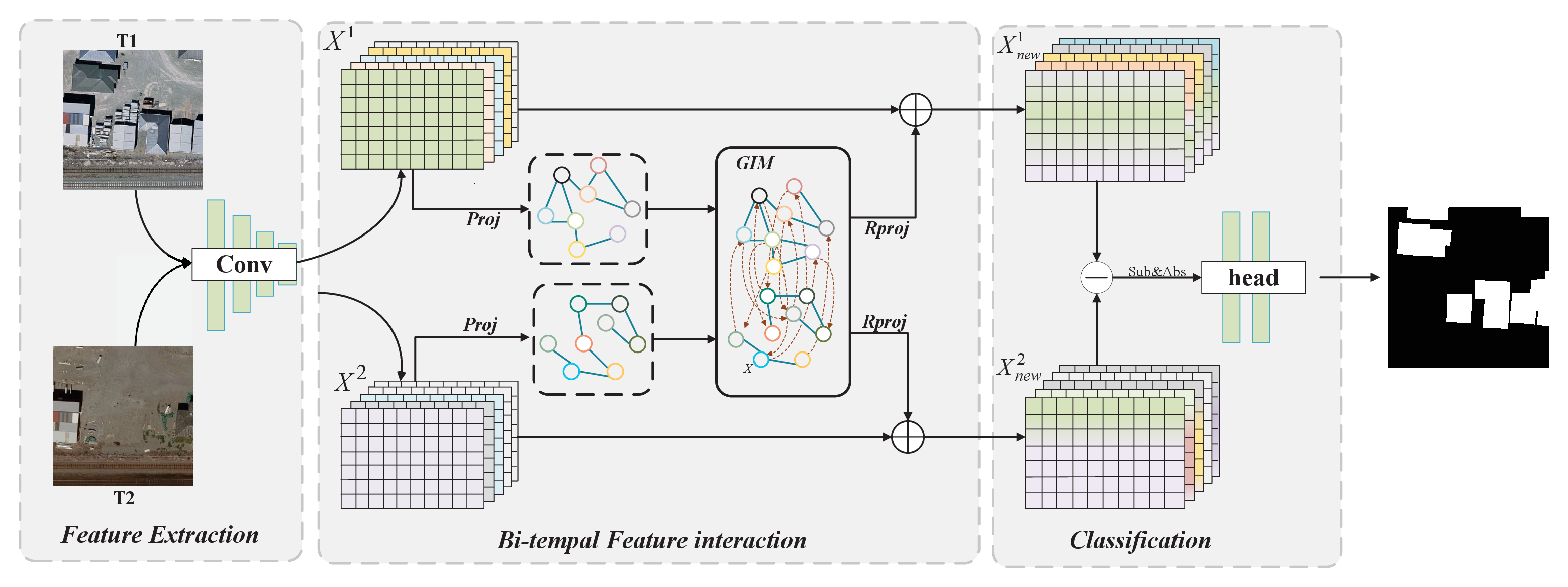}
		\caption{An overview of the proposed BGINet.}
		\label{figure_BGINet}%
	\end{figure*}

	\section{Proposed Method}
	\subsection{Overall Architecture}
	Figure 1 illustrates the architecture of the proposed BGINet. The network comprises two main components: a generic feature extractor and a Bitemporal Graph Interaction Module. To ensure model efficiency, we utilize the first three stages of ResNet-18 \cite{he2016deep} as the generic feature extractor. The Graph Interaction Module (GIM) branch takes the bitemporal generic features extracted by the feature extractor as input and maps them into a graph structure. Inspired by nonlocal operations, we incorporate self-attentiveness in the graph space to efficiently capture remote dependencies between bitemporal features. The evolved bitemporal features are then integrated with the original features. Finally, a 1 × 1 convolutional layer with a sigmoid activation function is applied to obtain the final difference map, which serves as an indicator of change.	
	\subsection{Graph Interaction Module (GIM)}
	Here, we detail the Graph Interaction Module (GIM). As shown in Figure 2, GIM are composed of three operations, namely, graph projection, 
    graph interaction, and graph reprojection.
	Given  bitemporal 2D feature map $X^1 \in \mathbb{R}^{h\times w \times c} $,$X^2 \in \mathbb{R}^{h\times w \times c}$
	, $\boldsymbol{x}_{i j}^{1}$,$\boldsymbol{x}_{i j}^{2} \in \mathbb{R}^d$ denotes 
	the \textit{d}-dimensional feature at $(i, j)$of bitemporal. The graph embedding can be denoted as
	$\mathcal{G}_1=(\mathcal{V}_1, \mathcal{Z}_1, \mathcal{A}_1)$ or $\mathcal{G}_2=(\mathcal{V}_2, \mathcal{Z}_2, 
	\mathcal{A}_2)$, where ${\mathcal{V}_1}$,${\mathcal{V}_2}$
	is a set of nodes, ${\mathcal{Z}_1}$,${\mathcal{Z}_2}$ is the feature matrix,
	and ${\mathcal{A}_1}$,${\mathcal{A}_2}$ is the affinity matrix between the nodes.

	\emph{1) Graph Projection}:
	To establish a correspondence between maps $X^1$ and $X^2$, we perform a feature mapping to obtain graphs
    $\mathcal{G}_1$ and $\mathcal{G}_2$. In this mapping, pixels with similar features are assigned to the same
    vertex in the graph. For simplicity, let's consider the feature mapping of the $t_1$ time phase as an example.
	Following the approach in \cite{2018Beyond}, we parameterize two matrices $\mathcal{W} \in \mathbb{R}^{|\mathcal{V}|\times d}$ and $\Sigma \in \mathbb{R}^{|\mathcal{V}| \times d}$, where $|\mathcal{V}|$ represents the number of vertices, which can be pre-specified.

    Each row $w_k$ of $\mathcal{W}$ serves as the anchor point for vertex $k$. To compute the soft assignment $q_{ij}^k$ of feature vector $\boldsymbol{x}_{ij}$ to $w_k$, we use the following equation:
	\begin{equation}
		q_{i j}^k=\frac{\exp \left(-\left\| \frac{\left(x_{i j}-w_k\right)} { \sigma_k} \right\|_2^2 / 2\right)}{\sum_k \exp \left(-\left\| \frac{\left(x_{i j}-w_k\right)} { \sigma_k} \right\|_2^2 / 2\right)}
	\end{equation}
	In this equation, $\sigma_k$ is the row vector of $\Sigma$ and is constrained to the range $(0, 1)$ using a sigmoid function. The numerator measures the similarity between the feature vector and the anchor point, while the denominator normalizes the assignment across all vertices.
	Next, we encode the vertex feature matrix $\mathcal{Z} \in R^{|\mathcal{V}| \times d}$ using the corresponding pixel features. For vertex $k$, we compute $z_k$, which represents the weighted average of the residuals between feature vectors $x_{ij}$ and $w_k$. Then, we normalize $z_k$ to obtain $z'_k$, the unit vector that forms the row vector of the feature matrix $\mathcal{Z}$ of graph $\mathcal{G}$:
	\begin{equation}
		z'_k=\dfrac{z_k}{\|z_k\|_2},\quad z_k=
			\left(\dfrac{\sum_{ij}q_{ij}^k\left(\boldsymbol{x}_{ij}-w_k\right)}{\sum_{ij}q_{ij}^k}\right)/\sigma_k
	\end{equation}
	Finally, the graph affinity matrix $\mathcal{A}$ is computed using the equation:

	\begin{equation}
		\mathcal{A}=\ \mathcal{ZZ}^\text{T}\in\mathbb{R}^{|\mathcal{V}|\times|\mathcal{V}|}
	\end{equation}
		
	\begin{figure*}[!t]
		\centering
		\vspace{-0.8cm}   
		\setlength{\abovecaptionskip}{0.cm} 
		\setlength{\belowdisplayskip}{1pt} 
		\includegraphics[scale=0.6]{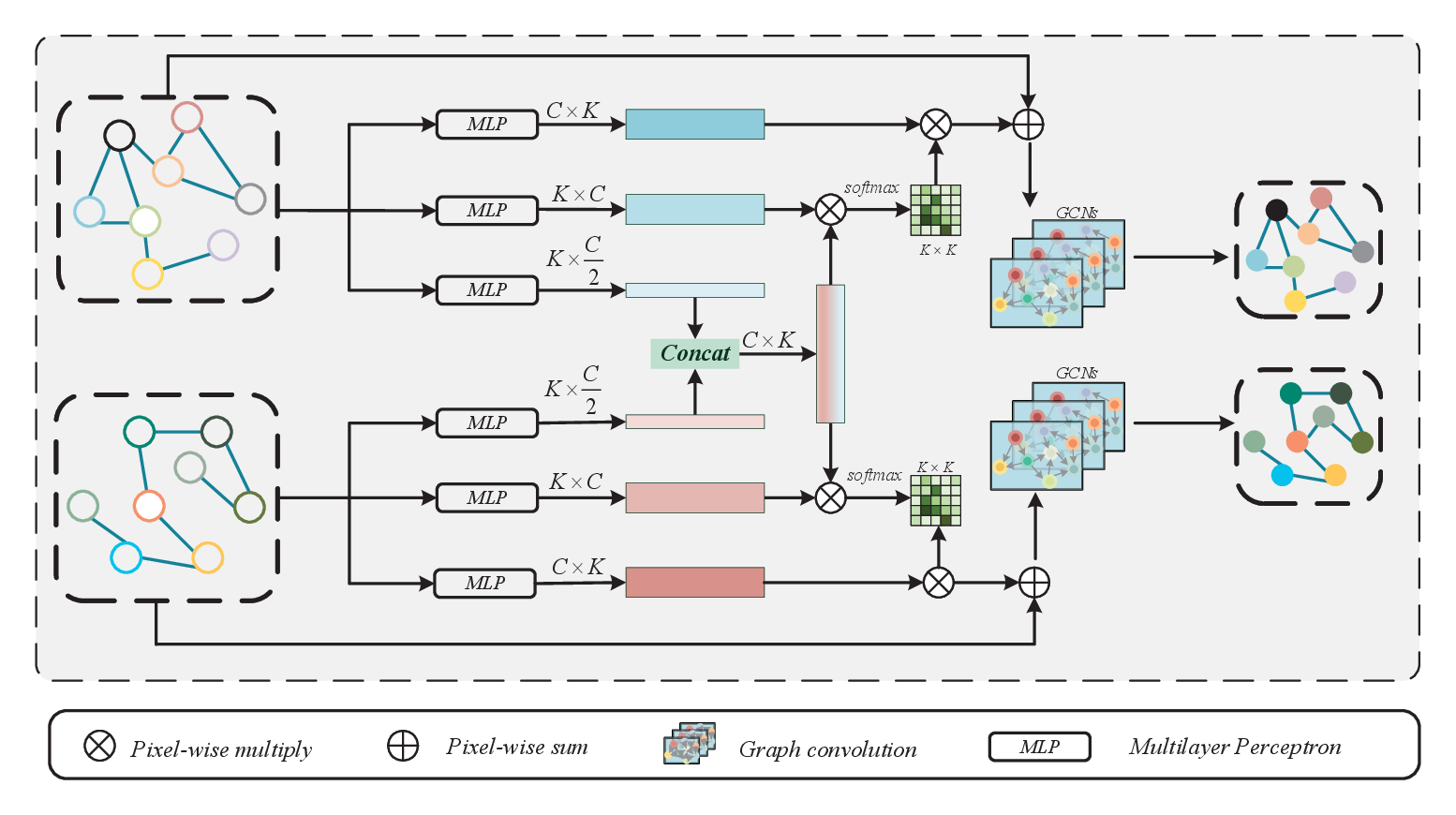}
		\caption{Graph Interaction Module (GIM).}
		\label{figure_BGINet}%
	\end{figure*}
	\emph{2) Graph Interaction}:
	The proposed GIM receiver map embeddings $\mathcal{G}_{1}=(\mathcal{V}_{1},\mathcal{Z}_{1},\mathcal{A}_{1})$, and $\mathcal{G}_{2}=(\mathcal{V}_{2},\mathcal{Z}_{2},\mathcal{A}_{2})$ as inputs form the feature mapping of $t_1$ time phase and $t_2$ time phase, respectively, 
	and models the between graph interaction and guides the inter-graph message passing from $V_1$ to $V_2$ and $V_2$ to $V_1$. 
	This goal leads us to take inspiration from non-local operations and DMINet [51] and compute inter-graph dependencies with a 
	concern mechanism. This operation of ours significantly reduces the number and computational complexity of parameters and achieves better results

	GIM models the betweengraph interaction and guides the inter-graph message passing from $\mathcal{Z}_1$ to $\mathcal{Z}_2$ or from $\mathcal{Z}_2$ to $\mathcal{Z}_1$
	As shown in Figure3,we use different multi-layer perceptrons (MLPs) to transform $\mathcal{Z}_1$ to the  query graph $\mathcal{Z}_{1}^q$,key graph $\mathcal{Z}_{1}^k$ and value graph $\mathcal{Z}_{1}^v$
	transform $\mathcal{Z}_2$ to the  query graph $\mathcal{Z}_{2}^q$,key graph $\mathcal{Z}_{2}^k$ and value $\mathcal{Z}_{2}^v$.Next, we unify $\mathcal{Z}_1^q$,$\mathcal{Z}_2^q$ as:
	\begin{equation}
		\mathcal{Z}^q=concat(\mathcal{Z}_1^q,\mathcal{Z}_2^q)
	\end{equation}
	Then, the similarity matrix $\mathcal{A}_{1 \rightarrow 2}^{\text {inter }}$ and $\mathcal{A}_{2 \rightarrow 1}^{\text {inter }}$$\in \mathbb{R}^{K \times K} $is calculated by a matrix multiplication as:
	\begin{align}
		\mathcal{A}_{1 \rightarrow 2}^{\text {inter }}=f_{\text {norm }}\left(\mathcal{Z}^{q \mathrm{T}} \times \mathcal{Z}_2^k\right) \\
		\mathcal{A}_{2 \rightarrow 1}^{\text {inter }}=f_{\text {norm }}\left(\mathcal{Z}^{q \mathrm{T}} \times \mathcal{Z}_1^k\right)
	\end{align}
	where $\mathcal{A}_{C \rightarrow E}^{\text {inter }} \in \mathbb{R}^{K \times K}$. After that, we can transfer semantic information from $\mathcal{Z}_1$ to $\mathcal{Z}_2$ and $\mathcal{Z}_2$ to $\mathcal{Z}_1$ by
	\begin{align}
		& \mathcal{Z}_1^{\prime}=f_{\text {GIM }}\left(\mathcal{Z}_2, \mathcal{Z}_1\right)=\left(\mathcal{A}_{2 \rightarrow 1}^{\text {inter }} \times \mathcal{Z}_1^{v \mathrm{T}}\right)+\mathcal{Z}_1 \\
	    & \mathcal{Z}_2^{\prime}=f_{\text {GIM }}\left(\mathcal{Z}_1, \mathcal{Z}_2\right)=\left(\mathcal{A}_{1 \rightarrow 2}^{\text {inter }} \times \mathcal{Z}_2^{v \mathrm{T}}\right)+\mathcal{Z}_2
	\end{align}
	After performing inter-graph interaction, we conduct the intra-graph reasoning by taking$\mathcal{Z}_1^{\prime}$ and $\mathcal{Z}_2^{\prime}$ as inputs to obtain enhanced graph representations.
	\begin{align}
		& \widetilde{\mathcal{Z}_1^{\prime}}=f_{\mathrm{GCN}}\left(\mathcal{Z}_1^\prime\right)=g\left(\mathcal{A}_{2 \rightarrow 1}^{\text {inter }} \mathcal{Z}_1^{\prime} W_1\right) \in \mathbb{R}^{C \times K} \\
		&  \widetilde{\mathcal{Z}_2^{\prime}}=f_{\mathrm{GCN}}\left(\mathcal{Z}_2^{\prime}\right)=g\left(\mathcal{A}_{1 \rightarrow 2}^{\text {inter }} \mathcal{Z}_2^{\prime} W_2\right) \in \mathbb{R}^{C \times K}
	\end{align}

	\emph{3) Graph Reprojection}: 
	To map the enhanced graph representations back to the original coordinate space, we revisit the assignments from the graph projection step.
	\begin{align}
		& X_1^{new}=Q_1  \widetilde{\mathcal{Z}_1^{\prime}}+X_1 \\
		& X_2^{new}=Q_2  \widetilde{\mathcal{Z}_2^{\prime}}+X_2
	\end{align}
	\section{Experiments and Results}
	\subsection{Experimental Dataset and Parameter Setting}
	The WHU Building Change Detection Dataset\cite{whu}:The data consists of two aerial images of two different time phases and the exact location, which contains 12796 buildings in 20.5 km2 with a resolution of 0.2 m and a size of 32570x15354. We crop the images to 256x256 size and randomly divide the training, validation, and test sets: 6096/762/762.

	Guangzhou Dataset(GZ-CD)\cite{GZ-CD}: The dataset was collected from 2006-2019, covering the suburbs of Guangzhou, China, and to facilitate the generation of image pairs, the Google Earth service of BIGEMAP software was used to collect 19 seasonally varying VHR image pairs with a spatial resolution of 0.55 m and a size range of 1006x1168 pixels to 4936x5224.We crop the images to 256x256 size and randomly divide the training, validation, and test sets: 2876/353/374.
	
	In the experiment, the number of vertices $|\mathcal{V}|$ is set to 32. We utilize the AdamW optimizer with weight decay 1e-4 and a polynomial schedule, where the initial learning rate is set to 0.0004. The total number of iterations is set to 100. The GPU used for the experiment is an NVIDIA V100.In this experiment, we employ the joint loss function consisting of Focal loss and Dice loss. For Focal loss, we set the parameters $\gamma$ and $\alpha$ to 2.0 and 0.2, respectively.The overall loss function is formulated as follows:
	\begin{equation}
		L_{t o t a l}=\lambda_1 \operatorname{Focal}(G T, \sigma(p))+\lambda_2 \operatorname{Dice}(G T, \sigma(p))
	\end{equation}
	Here, $\sigma$ represents the sigmoid activation function. We denote the model prediction as $p$, and $\lambda_1$ and $\lambda_2$ as the coefficients of the Focal loss and Dice loss, respectively. In this experiment, we set $\lambda_1$ and $\lambda_2$ to 0.5 and 1, respectively.
	\subsection{Experimental Results and Comparison}
	Our comparison experiments evaluate the trade-off between accuracy, number of parameters, and floating-point operations (FLOP). The quantitative results for the two datasets are presented in Table I and Table II, respectively.  The best-performing model in each column is highlighted in bold, while the second-best model is underlined. The tables provide a comprehensive view of the performance metrics, allowing us to analyze the accuracy and efficiency of different models	\begin{table}[h]
		\centering
		\setlength{\abovecaptionskip}{0.cm} 
		\renewcommand\arraystretch{1.16}
		\setlength{\tabcolsep}{7pt} 	
		\caption{QUANTITATIVE RESULTS OF THE CD METHODS ON THE WHU DATASET}
	\begin{tabular}{cccccc}
			\hline
			Model           & Precision (\%)            & Recall (\%)            & F1-score		        & Params (m)  \\
			\hline 
			FC-EF\cite{FC-EF}           &  91.19            & 85.30             & 88.15  		    & \textbf{1.1} \\
			FC-Siam-conc\cite{FC-EF}    & 69.04             & 84.93             & 76.17 	        & 1.55  \\	
			FC-Siam-diff\cite{FC-EF}    & 60.66             & \textbf{91.24}    & 72.87 		    & \underline{1.35} \\
			STANet\cite{stanet}          & 91.73 & 73.39             & 81.54  		    & 12.18\\
			SNUNet          & 75.23             & 89.12             & 81.59  		    & 12.04\\
			BIT             & 87.44             & \underline{90.24} & 88.82  	        & 3.55\\
			DMINet          & \textbf{93.84}    & 86.25             & \underline{88.69} & 6.24\\
			BGINEt          & \underline{91.84}             & 90.22             & \textbf{91.02}    &2.88 \\
			\hline
	\end{tabular}
	\end{table}
	\begin{table}[H]
		\centering 
		\setlength{\abovecaptionskip}{0.cm} 
		\renewcommand\arraystretch{1.16}
		\setlength{\tabcolsep}{7pt} 	
		\caption{QUANTITATIVE RESULTS OF THE CD METHODS ON THE GZ DATASET}
		\begin{tabular}{cccccc}
			\hline
			Model           & Precision (\%)            & Recall (\%)            & F1-score		        & Params (m)  \\
			\hline
			FC-EF\cite{FC-EF}           & 85.92             & 78.43              & 82.00  		     & \textbf{1.1} \\
			FC-Siam-conc\cite{FC-EF}    & 87.63             & 83.50              & 85.52 	         & 1.55  \\	
			FC-Siam-diff\cite{FC-EF}    & \textbf{90.47} & 79.45              & 84.60 		     & \underline{1.35} \\
			STANet\cite{stanet}          & 88.40    & 78.84              & 83.35  		     & 12.18\\
			SNUNet          &  \underline{89.61}             & 84.40              & 86.92  		     & 12.04\\
			BIT             & 86.38             & \textbf{88.60}     & \underline{87.48} & 3.55\\
			DMINet          &89.31             & 83.901             & 86.52  		     & 6.24\\
			BGINEt          & 88.52             & \underline{88.00}  & \textbf{88.25}    & 2.88 \\
			\hline
		\end{tabular}
	\end{table}
	\begin{figure*}[!t]
		\centering
		\vspace{-0.8cm}   
		\setlength{\abovecaptionskip}{0.1cm} 
		\setlength{\belowdisplayskip}{1pt} 
		\includegraphics[scale=0.27]{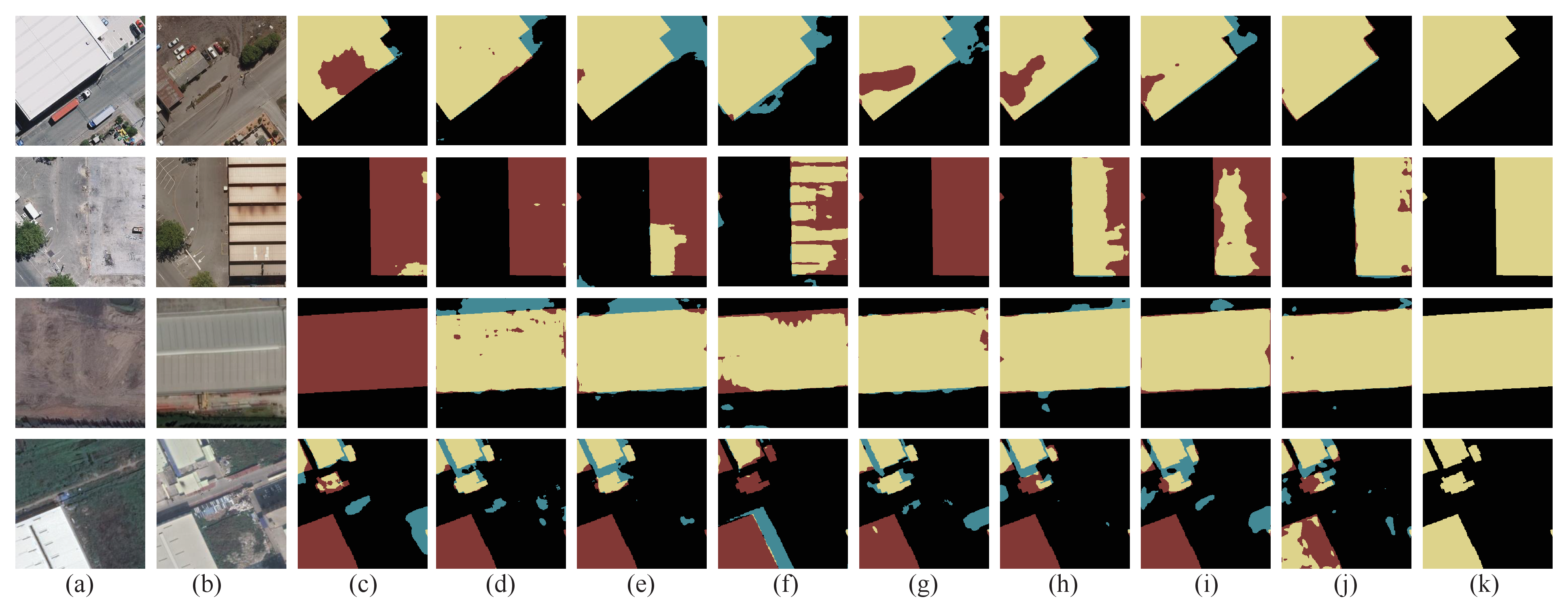}
		\caption{Figure 4 showcases the qualitative results of different methods on the GZ-CD and WHU-CD datasets. The top two rows correspond to the WHU-CD dataset, while the bottom two rows pertain to the GZ-CD dataset. Each subfigure represents a specific element, as follows: (a) RGB image of the first date, (b) RGB image of the second date, (c) FC-EF, (d) FC-Siam-conc, (e) FC-Siam-diff, (f) STANet, (g) SNUNet, (h) BIT, (i) DMINet, (j) BGINet, and (k) ground-truth labels. The true positive (TP) regions are highlighted in yellow, the false positive (FP) regions in red, and the false negative (FN) regions in blue. The true negative (TN) regions are depicted in black. For a more detailed observation, we recommend zooming in on the figure.}
		\label{figure_compare}%
	\end{figure*}
	In Fig. 3 and Fig. 4, we present a tradeoff analysis between the F1 score and computational cost for various classical remote sensing image change detection methods and recently proposed methods (e.g., STANet, BIT, DMINet, etc.). These figures provide valuable insights into different approaches' performance and computational requirements.The results indicate that while the aforementioned methods demonstrate good performance, they often have a significant computational overhead. On the other hand, baseline models like FC-EF require minimal computational resources but fall short in accuracy. Our proposed method, however, achieves a favorable balance between accuracy and computational overhead.By examining the figures, it is evident that our method outperforms the baseline models in terms of accuracy while still maintaining a manageable computational cost. This highlights the effectiveness and efficiency of our approach in remote sensing image change detection tasks.
	\begin{figure}[h]
		\vspace{-1em}
		\centering
		\setlength{\abovecaptionskip}{3pt} 
		\includegraphics[scale=0.5]{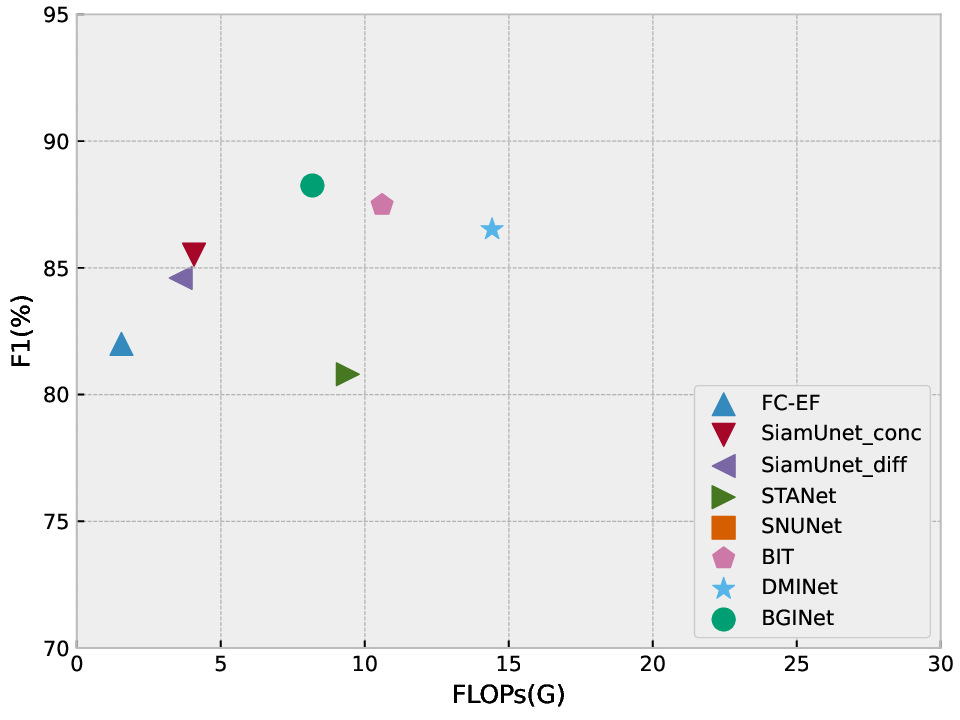}
		\caption{Comparing the complexity of different methods in terms of FlOP (computational cost) and F1-score on the GZ-CD dataset.}
		\label{figure_results22}%
	\end{figure}
	\begin{figure}[h]
		\vspace{-1.5em}
		\centering
		\setlength{\abovecaptionskip}{3pt} 
		\includegraphics[scale=0.5]{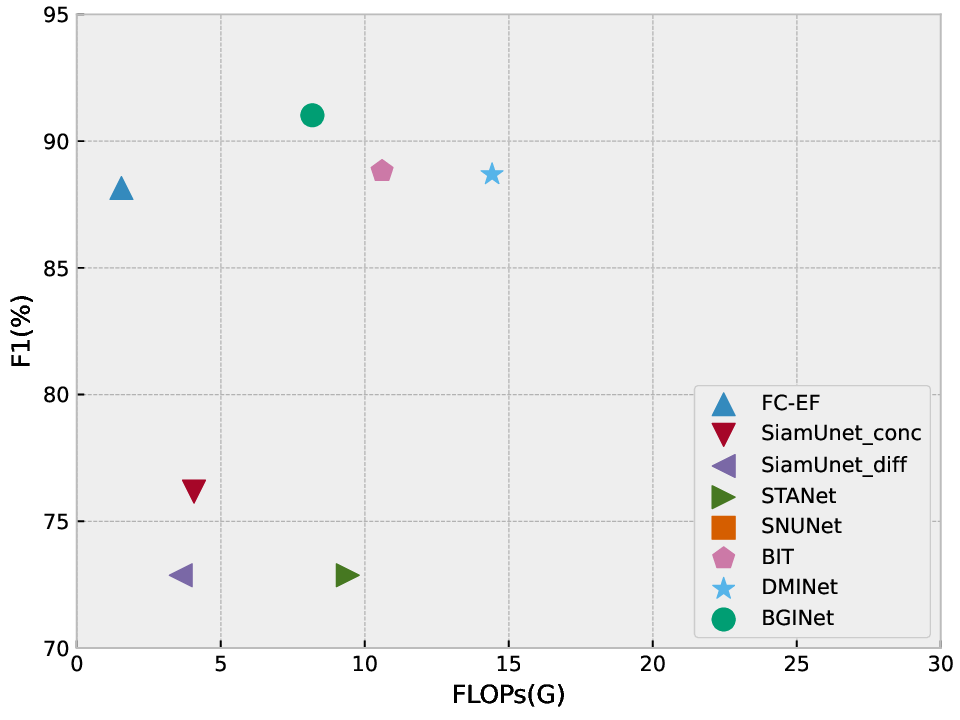}
		\caption{Comparing the complexity of different methods in terms of FlOP (computational cost) and F1-score on the WHU-CD dataset.}
		\label{WHU_FlOPs_results}%
	\end{figure}
	\begin{table}[h]
		\centering
		\setlength{\abovecaptionskip}{0.cm} 
		\renewcommand\arraystretch{1.16}
		\setlength{\tabcolsep}{7pt} 	
		\caption{QUANTITATIVE RESULTS OF THE CD METHODS ON THE WHU DATASET}
	\begin{tabular}{cccccc}
			\hline
			Model           & Precision (\%)            & Recall (\%)            & F1-score	 \\
			\hline 
			$Baseline_{gz}$		& 87.47             & 85.37             & 86.41  		     \\		
			$BGINEt_{gz}$        & 88.52             & 88.00             & \textbf{88.25}   \\
			$Baseline_{whu}$		& 85.70             & 92.20             & 88.80  		  \\
			$BGINEt_{whu}$          & 91.84             & 90.22             & \textbf{91.02}   \\
			\hline
	\end{tabular}
	\end{table}
	\subsection{Ablation Experiment}
	To validate the effectiveness of our proposed BGINet, we conducted ablation experiments on network knots using two publicly available datasets. The results of these experiments are presented in Table 3.
	As a baseline model, we selected resnet18 and utilized only the first three stages of the network. Upon introducing GIM (Graph Interaction Module) on the GZ-CD dataset, we observed improved precision, recall, and F1 scores by $1.05\%, 2.63\%, and 1.84\%$, respectively. On the WHU dataset, although there was a decrease in the recall by $2.02\%$, we observed improvements in precision and F1 scores by $6.14\%$ and $2.2\%$, respectively.
	These findings highlight the positive impact of our proposed BGINet, particularly when integrated with GIM, in enhancing the performance of remote sensing image change detection. The ablation experiments demonstrate the introduced graph interaction module's significance and contribution to improving precision, recall, and overall F1 score on both datasets.
	Overall, the results affirm the effectiveness of our approach and its potential for advancing the field of remote sensing image change detection.

	\section{Conclusion}
	In this letter, we introduce a novel method for improving change detection accuracy in dual-temporal images. By mapping the image features into the graph space and utilizing the Graph Interaction Module (GIM), we enable effective feature interaction and mitigate the influence of pseudo-change. Our proposed approach achieves a lightweight implementation, offering a favorable tradeoff between accuracy, number of parameters, and computational complexity. Experimental results on two publicly available datasets demonstrate the effectiveness of our model, with our approach achieving state-of-the-art performance on the GZ-CD dataset.


\begin{thebibliography}{1}		
		\bibitem{MSCANet}
		M.~Liu, Z.~Chai, H.~Deng, and R.~Liu, ``A cnn-transformer network with
		  multiscale context aggregation for fine-grained cropland change detection,''
		  \emph{IEEE Journal of Selected Topics in Applied Earth Observations and
		  Remote Sensing}, vol.~15, pp. 4297--4306, 2022,
		  doi:{10.1109/JSTARS.2022.3177235}.
		
		\bibitem{BIT}
		H.~Chen, Z.~Qi, and Z.~Shi, ``Remote sensing image change detection with
		  transformers,'' \emph{IEEE Transactions on Geoscience and Remote Sensing},
		  vol.~60, pp. 1--14, 2022, doi:{10.1109/TGRS.2021.3095166}.
		
		\bibitem{PCA}
		T.~Celik, ``Unsupervised change detection in satellite images using principal
		  component analysis and $k$-means clustering,'' \emph{IEEE Geoscience and
		  Remote Sensing Letters}, vol.~6, no.~4, pp. 772--776, 2009,
		  doi:{10.1109/LGRS.2009.2025059}.
		
		\bibitem{CVA}
		N.~Zerrouki, F.~Harrou, and Y.~Sun, ``Statistical monitoring of changes to land
		  cover,'' \emph{IEEE Geoscience and Remote Sensing Letters}, vol.~15, no.~6,
		  pp. 927--931, 2018, doi:{10.1109/LGRS.2018.2817522}.
		
		\bibitem{snunet}
		S.~Fang, K.~Li, J.~Shao, and Z.~Li, ``Snunet-cd: A densely connected siamese
		  network for change detection of vhr images,'' \emph{IEEE Geoscience and
		  Remote Sensing Letters}, vol.~19, pp. 1--5, 2022,
		  doi:{10.1109/LGRS.2021.3056416}.
		
		\bibitem{rdpnet}
		H.~Chen, F.~Pu, R.~Yang, R.~Tang, and X.~Xu, ``Rdp-net: Region detail
			preserving network for change detection,'' \emph{IEEE Transactions on
			Geoscience and Remote Sensing}, vol.~60, pp. 1--10, 2022,
			doi:{10.1109/TGRS.2022.3227098}.
		
		\bibitem{senet}
				J.~Hu, L.~Shen, and G.~Sun, ``Squeeze-and-excitation networks,'' in
				\emph{Proceedings of the IEEE conference on computer vision and pattern
				recognition}, 2018, pp. 7132--7141.
		
		\bibitem{cbam}
			S.~Woo, J.~Park, J.-Y. Lee, and I.~S. Kweon, ``Cbam: Convolutional block
			attention module,'' in \emph{Proceedings of the European conference on
			computer vision (ECCV)}, 2018, pp. 3--19.

		\bibitem{MSCANet}
			M.~Liu, Z.~Chai, H.~Deng, and R.~Liu, ``A cnn-transformer network with
				multiscale context aggregation for fine-grained cropland change detection,''
				\emph{IEEE Journal of Selected Topics in Applied Earth Observations and
				Remote Sensing}, vol.~15, pp. 4297--4306, 2022,
				doi:{10.1109/JSTARS.2022.3177235}.
			
		\bibitem{A1999Change}
			A.~Abuelgasim, W.~Ross, S.~Gopal, and C.~Woodcock, ``Change detection using
			adaptive fuzzy neural networks: Environmental damage assessment after the
			gulf war,'' \emph{Remote Sensing of Environment}, 1999.
			
		\bibitem{frick2019framework}
			A.~Frick and S.~Tervooren, ``A framework for the long-term monitoring of urban
			green volume based on multi-temporal and multi-sensoral remote sensing
			data,'' \emph{Journal of geovisualization and spatial analysis}, vol.~3,
			no.~1, p.~6, 2019.
		  
		\bibitem{vaswani2017attention}
			A.~Vaswani, N.~Shazeer, N.~Parmar, J.~Uszkoreit, L.~Jones, A.~N. Gomez,
			{\L}.~Kaiser, and I.~Polosukhin, ``Attention is all you need,''
			\emph{Advances in neural information processing systems}, vol.~30, 2017.
		  
		\bibitem{DMINet}
			Y.~Feng, J.~Jiang, H.~Xu, and J.~Zheng, ``Change detection on remote sensing
			images using dual-branch multilevel intertemporal network,'' \emph{IEEE
			Transactions on Geoscience and Remote Sensing}, vol.~61, pp. 1--15, 2023.

		\bibitem{2018Beyond}
			Y.~Li and A.~Gupta, ``Beyond grids: Learning graph representations for visual
			recognition,'' in \emph{Neural Information Processing Systems}, 2018.

		\bibitem{he2016deep}
			K.~He, X.~Zhang, S.~Ren, and J.~Sun, ``Deep residual learning for image
			recognition,'' in \emph{Proceedings of the IEEE conference on computer vision
			and pattern recognition}, 2016, pp. 770--778.

		\bibitem{whu}
			S.~Ji, S.~Wei, and M.~Lu, ``Fully convolutional networks for multisource
			building extraction from an open aerial and satellite imagery data set,''
			\emph{IEEE Transactions on geoscience and remote sensing}, vol.~57, no.~1,
			pp. 574--586, 2018.
		\bibitem{GZ-CD}
			D.~Peng, L.~Bruzzone, Y.~Zhang, H.~Guan, H.~Ding, and X.~Huang, ``Semicdnet: A
			  semisupervised convolutional neural network for change detection in high
			  resolution remote-sensing images,'' \emph{IEEE Transactions on Geoscience and
			  Remote Sensing}, vol.~59, no.~7, pp. 5891--5906, 2021,
			  doi:{10.1109/TGRS.2020.3011913}.

		\bibitem{stanet}
			H.~Chen and Z.~Shi, ``A spatial-temporal attention-based method and a new
			dataset for remote sensing image change detection,'' \emph{Remote Sensing},
			vol.~12, no.~10, 2020, doi:{10.3390/rs12101662}. [Online]. Available:
			\url{https://www.mdpi.com/2072-4292/12/10/1662}

		\bibitem{FC-EF}
			R.~Caye~Daudt, B.~Le~Saux, and A.~Boulch, ``Fully convolutional siamese
			  networks for change detection,'' in \emph{2018 25th IEEE International
			  Conference on Image Processing (ICIP)}, 2018, pp. 4063--4067,
			  doi:{10.1109/ICIP.2018.8451652}.
	\end{thebibliography}
\end{document}